\documentclass[conference]{IEEEtran}
\pdfoutput=1

\usepackage[T1]{fontenc}

\newif\ifinternalshare
\internalsharefalse          

\ifinternalshare
  \usepackage{draftwatermark}
  \SetWatermarkText{Pending Publication - Do Not Share}
  \SetWatermarkAngle{45}
  \SetWatermarkScale{0.3}
  \SetWatermarkColor[gray]{0.75}
\fi
\IEEEoverridecommandlockouts
\usepackage{cite}
\usepackage{mdframed}
\usepackage{amsmath,amssymb,amsfonts}
\usepackage{algorithm}
\usepackage{algorithmic}
\usepackage{graphicx}
\usepackage{textcomp}
\usepackage{xcolor}
\usepackage{booktabs}
\usepackage{array}
\usepackage{tcolorbox}
\tcbuselibrary{listings}
\tcbuselibrary{skins}
\tcbuselibrary{breakable}

\newtcolorbox{promptbox}{
  colback=gray!5,
  colframe=gray!40,
  boxrule=0.4pt,
  arc=1mm,
  fontupper=\scriptsize\ttfamily,
  left=2mm,
  right=2mm,
  top=1mm,
  bottom=1mm,
  listing only,
  listing options={
    basicstyle=\scriptsize\ttfamily,
    breaklines=true,
    breakatwhitespace=true,
    columns=fullflexible,
    keepspaces=true,
    upquote=true
  },
  enhanced,
  sharp corners,
  breakable
}
\def\BibTeX{{\rm B\kern  .05em{\sc i\kern  .025em b}\kern  .08em
    T\kern  .1667em\lower.7ex\hbox{E}\kern  .125emX}}

\usepackage[hidelinks]{hyperref}
    
\usepackage[protrusion=true,expansion=true]{microtype} 
\begin{document}

\title{AVATAAR: Agentic Video Answering via Temporal Adaptive Alignment and Reasoning}

\IEEEspecialpapernotice{Accepted in the 5th IEEE Big Data Workshop on Multimodal AI (MMAI 2025), Preprint Copy\\ Dec 8-11, Macau, China, 2025}


\author{\IEEEauthorblockN{Fang-Chun Yeh}
\IEEEauthorblockA{\textit{Ratings Data Science} \\
\textit{S\&P Global}\\
New York, USA \\
jessie.yeh@spglobal.com}
\and
\IEEEauthorblockN{Urjitkumar Patel}
\IEEEauthorblockA{\textit{Ratings Data Science} \\
\textit{S\&P Global}\\
New York, USA \\
urjitkumar.patel@spglobal.com}
\and
\IEEEauthorblockN{Chinmay Gondhalekar}
\IEEEauthorblockA{\textit{Ratings Data Science} \\
\textit{S\&P Global}\\
New York, USA \\
chinmay.gondhalekar@spglobal.com}
}

\maketitle

\begin{abstract}

With the increasing prevalence of video content, effectively understanding and answering questions about long form videos has become essential for numerous applications. Although large vision language models (LVLMs) have enhanced performance, they often face challenges with nuanced queries that demand both a comprehensive understanding and detailed analysis. To overcome these obstacles, we introduce AVATAAR, a modular and interpretable framework that combines global and local video context, along with a Pre Retrieval Thinking Agent and a Rethink Module. AVATAAR creates a persistent global summary and establishes a feedback loop between the Rethink Module and the Pre Retrieval Thinking Agent, allowing the system to refine its retrieval strategies based on partial answers and replicate human-like iterative reasoning. On the CinePile benchmark, AVATAAR demonstrates significant improvements over a baseline, achieving relative gains of +5.6\% in temporal reasoning, +5\% in technical queries, +8\% in theme-based questions, and +8.2\% in narrative comprehension. Our experiments confirm that each module contributes positively to the overall performance, with the feedback loop being crucial for adaptability. These findings highlight AVATAAR's effectiveness in enhancing video understanding capabilities. Ultimately, AVATAAR presents a scalable solution for long-form Video Question Answering (QA), merging accuracy, interpretability, and extensibility.

\end{abstract}


\begin{figure*}[t]
\centering
\includegraphics[width=0.9\textwidth]{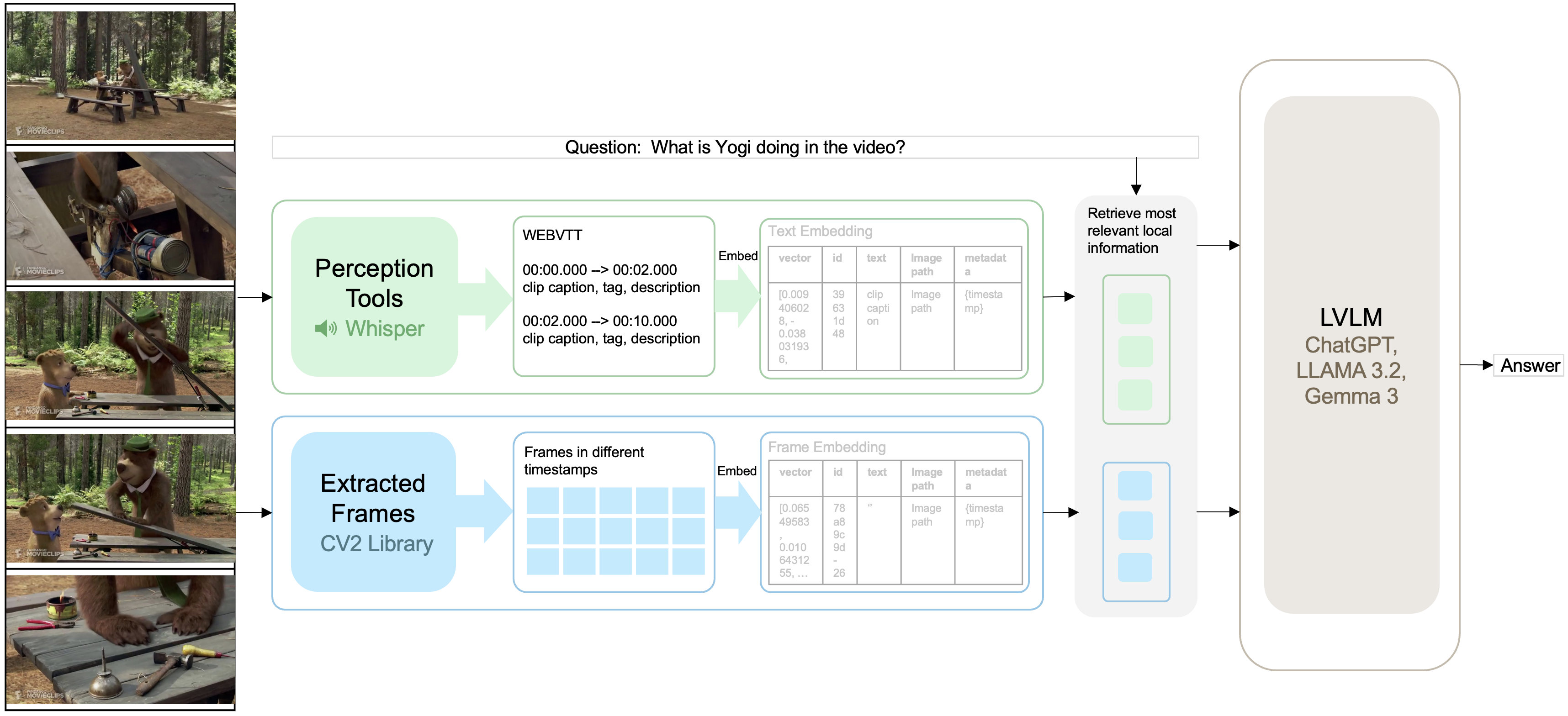} 
\caption{Traditional video question answering pipeline}
\label{fig:framework_traditional}
\end{figure*}

\section{Introduction}

Video is a crucial medium for sharing information, supporting a wide range of applications from online education and entertainment to surveillance and autonomous systems. Consequently, video question answering (VQA) has become a key challenge in artificial intelligence, requiring systems to interpret complex visual and textual cues that are spread across time \cite{zhu2020comprehensivestudydeepvideo}.

Recent work has explored various strategies to advance video understanding, including frame aggregation for improved action recognition, multimodal embedding models for better vision language alignment, and video augmentation techniques. Multi-agent architectures have also been proposed \cite{VideoMultiAgents}, where specialized modules for different modalities collaborate to generate answers. Another promising direction involves training larger foundation models \cite{gemini25}, although these approaches are often computationally intensive and challenging to scale \cite{scalability_issues}. 

Despite this progress, many existing systems continue to struggle with limited context windows, a lack of persistent memory, and static retrieval mechanisms. These constraints hinder their ability to handle complex, multi step reasoning over long videos. To address these challenges, we introduce AVATAAR, a flexible framework that integrates global video summaries, dynamic query refinement via a Pre Retrieval Thinking Agent, and iterative feedback through a Rethink Module. This design enables more adaptable, context aware inference across diverse question types and domains, going beyond static architectures and fixed retrieval strategies.

\begin{figure*}[t]
\centering
\includegraphics[width=0.9\textwidth]{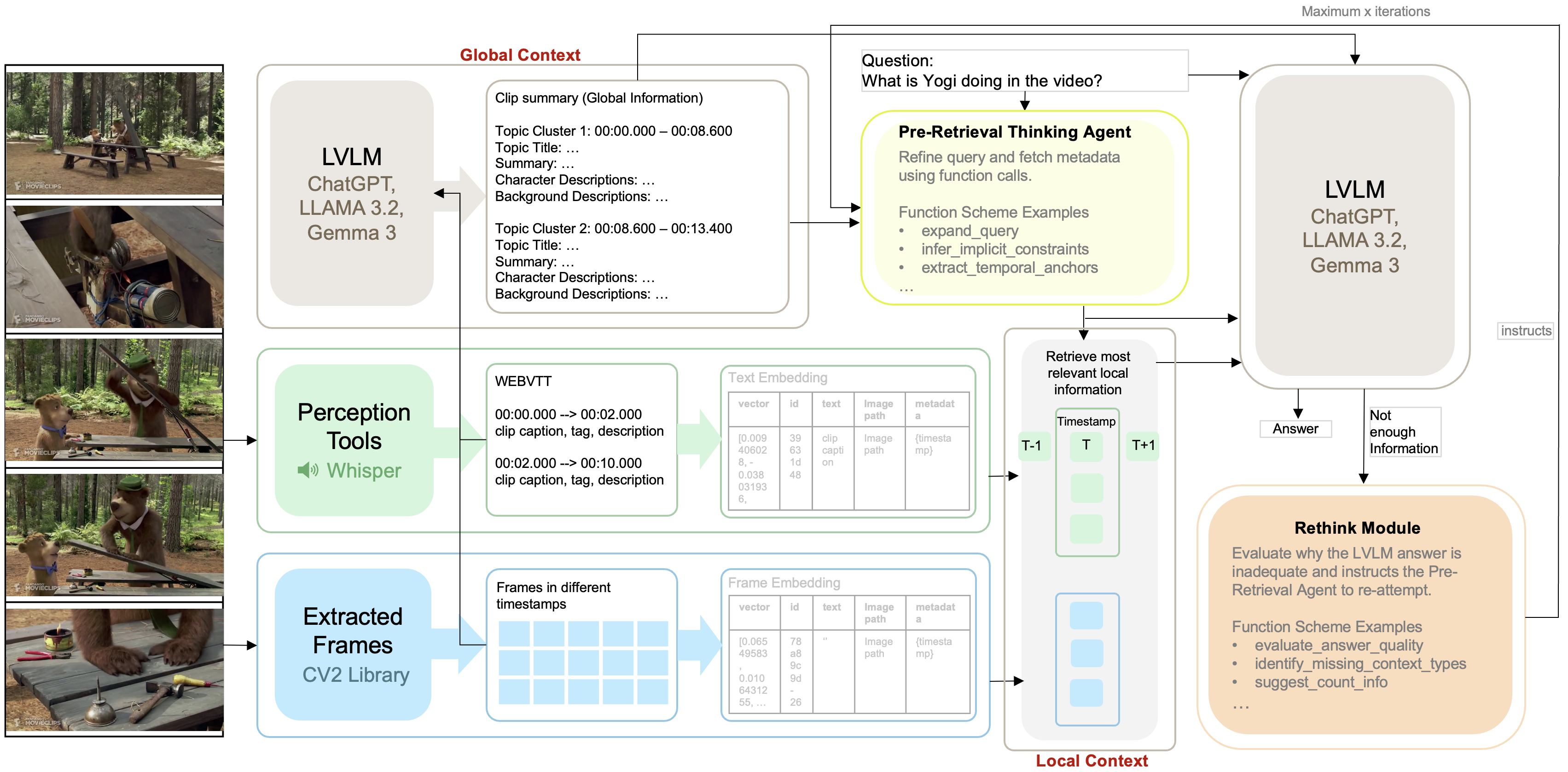} 
\caption{AVATAAR: Think, Retrieve, Rethink - Agentic Video QA Framework}
\label{fig:framework_AVATAAR}
\end{figure*}

As shown in Figure~\ref{fig:framework_AVATAAR}, AVATAAR starts by breaking the video into segments that are easy to manage, keeping in mind the context window limits of LVLMs. Next, it builds a global summary that highlights the main storyline, important events, and interactions between characters throughout the entire video. This summary acts as a lasting memory, helping to guide the reasoning process that follows.

When a user poses a question, the Pre Retrieval Thinking Agent leverages the global summary, along with prior feedback from the Rethink Module (if available), to determine the appropriate reasoning steps, including triggering function calls. It then refines and expands the query by injecting relevant context and disambiguating intent, ensuring that retrieval focuses on the most salient local segments, specific frames, and transcripts aligned with the user’s informational needs.

If the LVLM’s initial response is incomplete or ambiguous, the Rethink Module re-analyzes the question, global context, retrieved evidence, and the generated answer to diagnose potential gaps. It then issues targeted feedback to the Pre Retrieval Thinking Agent, prompting further query refinement or an updated retrieval strategy. This feedback loop continues for a fixed number of iterations or until a satisfactory answer is produced, mimicking human-like reasoning through progressive, context aware inference.

We evaluate AVATAAR on CinePile \cite{rawal2024cinepile}, demonstrating that our architectural and data processing enhancements substantially improve performance on challenging video QA tasks, especially those requiring nuanced spatio-temporal reasoning and multi-step evidence integration.

Our main contributions are as follows: 
\begin{itemize}
  \item We present a modular framework for long-form video QA that combines global summarization, local retrieval, and iterative reasoning to improve accuracy and interpretability over a traditional video chat baseline.
  \item We describe a dynamic summary generation module using LVLMs, which adaptively selects frames and transcriptions to create a persistent global context for video QA.
  \item A Pre Retrieval Thinking Agent that refines user queries using global context and prior feedback, enabling more precise retrieval, especially for complex, entity-based, or quantitative questions that are difficult for static RAG pipelines.
  \item A Rethink Module that diagnoses gaps in earlier answers and issues targeted instructions to guide the Pre Retrieval Agent for the next round of retrieval and reasoning, yielding a simple form of agentic RAG.
  \item We show that incorporating timestamp information into retrieval and reasoning modules significantly improves performance on temporally grounded questions on the CinePile benchmark, and we discuss the computational and latency trade-offs introduced by AVATAAR's iterative design.
\end{itemize}

\section{Related Work}

\subsection{Multimodal Embedding Models}

Aligned vision language spaces are central to contemporary video QA. Dual encoder foundations such as CLIP \cite{clip} and ALIGN \cite{align} project images (or individual frames) and text into a shared latent space, making them natural backbones for this task. BridgeTower \cite{bridgetower} generalizes this approach by inserting bridge layers that couple intermediate features from pretrained encoders using CLIP for the visual stream and RoBERTa \cite{liu2019roberta} for the language stream to enable fine grained semantic correspondence. In video QA practice, these models are adapted by temporally aggregating per frame embeddings and aligning them with transcripts or with shot level segments.

\subsection{Large Vision  Language Models for Video}

Large Vision Language Models (LVLMs) have been pivotal in advancing multimodal reasoning for video understanding tasks. Early approaches such as Flamingo \cite{flamingo} and BLIP  2 \cite{blip2} were among the first to leverage vision language pretraining for open-ended video question answering and temporal comprehension. Building on this foundation, instruction tuned models like InternVideo \cite{internvideo}, Video LLaMA \cite{videollama}, and Video LLaVA \cite{videollava} have demonstrated improved performance in event recognition and detailed visual text alignment by densely associating frames and transcripts.

The field has recently seen the emergence of more general purpose and proprietary LVLMs from leading research organizations. Models such as GPT 4o from OpenAI \cite{gpt4o}, Gemini 2.5 from Google DeepMind \cite{gemini25}, Claude 3 Opus from Anthropic \cite{claude3}, and Meta’s LLaMA 3 \cite{grattafiori2024llama3} and Chameleon \cite{chameleonteam2025chameleonmixedmodalearlyfusionfoundation} extend multimodal reasoning capabilities to the unified processing of images, videos, and audio.

Nevertheless, the majority of LVLMs continue to face challenges, such as fixed input context windows and the absence of mechanisms for dynamic context management or iterative reasoning, which can hinder their effectiveness on long form video QA.




\subsection{Retrieval Augmented and Agentic Reasoning}

Retrieval Augmented Generation (RAG) anchors generated answers in evidence retrieved from a corpus \cite{lewis2020rag, karpukhin2020dpr}. Classic RAG pipelines are single shot: a fixed query is issued once to a fixed retriever. Agentic variants depart from this by introducing multi-turn planning and self reflection. Reflexion \cite{shinn2023reflexion} equips agents to critique prior outputs and revise them through memory updates. A recent survey \cite{singh2025agenticrag} systematizes this design space, emphasizing tool use, iterative query reformulation, and modular planning capabilities crucial for ambiguous, multi step problems. Modular RAG frameworks have also been explored in domain specific settings such as cyber\cite{patel2024canal} and finance\cite{patel2024fanal, gondhalekar2025multifinrag}. 

In video QA, retrieval must respect the spatio-temporal structure. VideoRAG \cite{videorag} fuses structured concept graphs with visual descriptors to enable frame level lookup. Hierarchical alignment methods (e.g., Dig into Multi modal Cues \cite{ijcai2021-154}) tailor layered alignment to strengthen video-to-text retrieval. VideoMultiAgents \cite{VideoMultiAgents} further distributes reasoning across modality  specialist agents under a centralized coordinator.

Despite these advances, many systems still lack long lived memory, run-time adaptive retrieval policies, and feedback driven refinement. AVATAAR closes these gaps by combining a global video summary, a Pre Retrieval Thinking Agent, and a Rethink Module within an iterative reasoning loop, yielding more context aware, human-like inference on long-form video content.

\section{Methods}

The frameworks for video question answering in this study: a baseline pipeline that uses traditional RAG, and AVATAAR, our advanced agentic architecture that integrates global and local context, dynamic query refinement, and iterative reasoning. Below, we describe the methodology and components of both frameworks in detail.

\renewcommand{\arraystretch}{1.1} 

\begin{table*}[t]
\centering
\small
\caption{Function Schema for the Pre  Retrieval Agent}
\label{tab:function_schema_for_the_pre_retrieval_agent}
\begin{tabular}{
    p{4cm} | p{12cm}
}
\toprule
\textbf{Function Name} & \textbf{Purpose} \\
\midrule
expand\_query                  & Adds missing details, such as character descriptions, to enrich the query. \\
extract\_temporal\_anchors     & Returns timestamps of retrieved frames or estimates time  based cues from the video summary. \\
term\_frequency                & Counts how many times a term shows up in the transcription. \\
get\_action\_before\_an\_event & Suggests actions by inputting six consecutive frames before the event timestamp to the LVLMs. \\
get\_action\_after\_an\_event  & Suggests actions by inputting six consecutive frames after the event timestamp to the LVLMs. \\
web\_search  & Search web for unfamiliar terms in user query. \\
multi\_hop\_query\_generation & Break down a complex question into a sequence of simpler sub  queries, retrieve for each, and aggregate the results. \\
\bottomrule
\end{tabular}
\end{table*}

\subsection{Baseline Framework}

As depicted in Figure \ref{fig:framework_traditional}, the basic framework for video question answering consists of three primary stages: data preprocessing, retrieval, and answer generation.

\vspace{2mm}
\subsubsection{Data Pre-processing}

We pre-process raw videos with automated Python tooling. Whisper \cite{radford2022whisper} produces a time stamped sequence of transcriptions ${T_j}$, where each $T_j$ corresponds to segment $j$. In parallel, OpenCV (CV2) samples frames every $y$ seconds to obtain ${F_i}$. Temporal alignment of these two streams yields synchronized visual and textual signals, enabling coherent analysis and remaining robust even when audio tracks are absent.

\vspace{2mm}
\vspace{2mm}

\subsubsection{Embedding and Retrieval}

Video frames and transcript segments are encoded into a shared $z$-dimensional space using a CLIP vision encoder and a compatible text encoder.
Each video frame $F_i$ and transcript segment $T_j$ is mapped into the same $z$-dimensional embedding space:
\begin{equation}
\begin{split}
\mathrm{Emb}_{f_i} &= \mathrm{Embed}_\mathrm{img}(F_i) \in \mathbb{R}^z, \\
\mathrm{Emb}_{t_j} &= \mathrm{Embed}_\mathrm{txt}(T_j) \in \mathbb{R}^z, \\
\mathrm{Emb}_q &= \mathrm{Embed}_\mathrm{txt}(q) \in \mathbb{R}^z
\end{split}
\end{equation}
where $\mathrm{Emb}_{f_i}$, $\mathrm{Emb}_{t_j}$, and $\mathrm{Emb}_q$ denote the embeddings of frames, transcripts, and the user query, respectively.

To retrieve the most relevant frames, we compute the cosine similarity between the query embedding $\mathrm{Emb}_q$ and each frame embedding $\mathrm{Emb}_{f_i}$:
\begin{equation}
\mathrm{sim}(\mathrm{Emb}_q, \mathrm{Emb}_{f_i}) = \frac{\mathrm{Emb}_q \cdot \mathrm{Emb}_{f_i}}{\|\mathrm{Emb}_q\| \|\mathrm{Emb}_{f_i}\|}
\end{equation}
The indices of the top  $n$ relevant frames are:
\begin{equation}
I^\mathrm{frame}_{\mathrm{top}\text{  }n} = \underset{I \subset \{1,\ldots,N\}, |I|=n}{\arg\max} \sum_{i \in I} \mathrm{sim}(\mathrm{Emb}_q, \mathrm{Emb}_{f_i})
\end{equation}

Similarly, for transcriptions, we compute:
\begin{equation}
\mathrm{sim}(\mathrm{Emb}_q, \mathrm{Emb}_{t_j}) = \frac{\mathrm{Emb}_q \cdot \mathrm{Emb}_{t_j}}{\|\mathrm{Emb}_q\| \|\mathrm{Emb}_{t_j}\|}
\end{equation}
and select the indices of the top  $n$ relevant transcriptions:
\begin{equation}
J^\mathrm{trans}_{\mathrm{top}\text{  }n} = \underset{J \subset \{1,\ldots,M\}, |J|=n}{\arg\max} \sum_{j \in J} \mathrm{sim}(\mathrm{Emb}_q, \mathrm{Emb}_{t_j})
\end{equation}

The retrieved frames and transcriptions are then:
\begin{equation}
\{ F_i \mid i \in I^\mathrm{frame}_{\mathrm{top}\text{  }n} \}, \qquad
\{ T_j \mid j \in J^\mathrm{trans}_{\mathrm{top}\text{  }n} \}
\end{equation}

\vspace{2mm}
\subsubsection{Answer Generation}
The selected image  transcription pair $(F_{i^*}, T_{j^*})$, together with the user query $q$, is then passed to an LVLM, denoted as function $A$, to generate the answer:
\begin{equation}
\mathrm{Answer} = A(F_{i^*}, T_{j^*}, q)
\end{equation}
where $A(\cdot)$ denotes the LVLM’s answer generation operator, which leverages both the visual and textual context to produce a natural language response.

While this baseline approach effectively integrates multimodal information for basic video QA tasks, it is limited by its reliance on static retrieval and shallow context aggregation. These constraints motivate the development of more advanced architectures, as described in subsequent sections.

\subsection{AVATAAR Framework}

Building on conventional video QA pipelines, we introduce AVATAAR, a framework that explicitly models a think–retrieve loop and triggers a rethinking stage when an initial answer is inadequate. As illustrated in Figure~\ref{fig:framework_AVATAAR}, AVATAAR comprises three core components:
\vspace{2mm}

\begin{enumerate}
\item Global and Local Context Assisted Retrieval
\item Pre  Retrieval Thinking Agent
\item Rethink Module
\end{enumerate}
\vspace{2mm}

The overall architecture is shown in Figure~\ref{fig:framework_AVATAAR}; we describe each component in detail below.
\vspace{2mm}


\subsubsection{Global and Local context Assisted Retrieval}

In AVATAAR, we explicitly separate \emph{global} and \emph{local} context, each serving a distinct role in video understanding. The global context is a video level summary computed once per video and cached as a persistent reference. It captures overarching themes, principal characters, and salient background details, providing a stable foundation that remains consistent across all subsequent user queries.

By contrast, the local context is assembled on demand for each question. It consists of the most relevant frames and transcript spans retrieved conditionally on the query, and therefore changes dynamically from one query to the next. This query tailored evidence focuses the system on precisely the information needed to answer accurately, while the global summary anchors interpretation and maintains coherence across interactions.

\renewcommand{\arraystretch}{1.1} 

\begin{table*}[t]
\centering
\small
\caption{Question Types in the CinePile Dataset \cite{rawal2024cinepile}}
\label{tab:question  types}
\begin{tabular}{
    >{\centering\arraybackslash}p{0.5cm} 
    | p{3cm} 
    | p{6.5cm} 
    | p{4.3cm} 
    | >{\raggedleft\arraybackslash}p{1cm}
}
\toprule
\textbf{\#} & \textbf{Question Type} & \textbf{Description} & \textbf{Example} & \textbf{Count} \\
\midrule
1 & Character and Relationship Dynamics  & Templates about characters’ actions, motivations, and relationships. & How are Darren and Reed related to each other? & 2068 \\
2 & Narrative and Plot Analysis          & Templates focused on the movie’s storyline, plot twists, and narrative structure. & What does Peter notice about Edgar during their interaction? & 463 \\
3 & Setting and Technical Analysis       & Templates about the movie’s setting, environment, technical aspects, and object usage. & What does Reed do after opening the door? & 1521 \\
4 & Temporal                             & Templates assessing understanding of a movie’s timing and sequence of events. & How many times does Darren say "Ow!"? & 683 \\
5 & Theme Exploration                    & Templates about the movie’s themes, symbols, motifs, subtext, and emotional or moral elements. & How does the emotional tone transition during the scene? & 189 \\
\bottomrule
\end{tabular}
\end{table*}

\vspace{2mm}

\textbf{Dynamic Summary Generation (Global Context):}

Leveraging LVLMs, we generate a comprehensive summary of the video by aligning extracted frames with transcriptions, clustering content by topic, and describing key characters and backgrounds. This summary encapsulates the global context and serves as a persistent reference for all downstream queries, as depicted in the upper section of Figure \ref{fig:framework_AVATAAR}.

To accommodate the limited context window of the LVLM, we partition the video’s transcriptions and corresponding frames into batches, where each batch fits within the model’s maximum input capacity. For each batch, the LVLM generates a partial summary, including topic clusters, character descriptions, and background information. These partial summaries are then aggregated to form the global summary for the entire video. This batching strategy ensures that even for long videos exceeding the LVLM’s native context window, the framework can systematically process all available content without information loss, maintaining coherence across the global summary.

The prompt template used to generate the summary for each batch is as follows:

\begin{promptbox}
You are provided with a transcript in WEBVTT format and a set of frames extracted from the corresponding video. Please perform the following tasks:

\vspace{2mm}
\textbf{Topic Clustering}:
Analyze the captions and cluster them into coherent topics.

For each cluster, provide:\\
   The start and end timestamps covered by the topic.\\
   The \texttt{start\_time} of each topic cluster must be exactly equal to the \texttt{end\_time} of the previous topic cluster, except for the first cluster. \\
   A brief topic title and a short summary.

\vspace{2mm}

\textbf{Character Description}:
Identify any characters mentioned or appearing in each topic cluster.
For each character, provide a brief description based on both the transcript and the video frames (e.g., appearance, attire, actions, emotions).
Ensure that the character’s name appears in the topic title or summary for clusters where they are relevant.

\vspace{2mm}

\textbf{Frame References}:
For each topic cluster, reference the specific video frames that correspond to the start and end timestamps.
If a character is described, refer to the frame(s) where their appearance or actions are most clearly depicted.

\vspace{2mm}

\textbf{Output Format:} For each topic cluster, include: \\
   Start timestamp \\
   End timestamp \\
   Topic Title (include character names if relevant) \\
   Short summary of the topic \\
   Character Descriptions \\
   Background Descriptions \\
   Referenced Video Frames
\end{promptbox}

\vspace{2mm}

\subsubsection{Pre  Retrieval Thinking Agent}

The Pre  Retrieval Thinking Agent is an agent that receives three types of input: the user query $q$, the global context summary, and instructions from the Rethink Module. Leveraging these inputs, the agent examines the user query alongside the descriptions of available functions $\{f_1, f_2, \ldots, f_n\}$, and, using its own language understanding capabilities, dynamically selects and applies the most appropriate specialized functions to refine the query.

Let $c$ denote the set of contextual information or outputs fetched by the selected function callings. The refined query $q_r$ is constructed by augmenting the original query $q$ with $c$:
\begin{equation}
q_r = \mathrm{Refine}(q, c)
\end{equation}
where $\mathrm{Refine}(\cdot)$ denotes the process of incorporating additional context or information into the original query.

The refined query $q_r$ is then embedded as:
\begin{equation}
\mathrm{Emb}_{q_r} = \mathrm{Embed}_\mathrm{txt}(q_r) \in \mathbb{R}^z
\end{equation}

Given sets of frame embeddings $\{\mathrm{Emb}_{f_i}\}$ and transcription embeddings $\{\mathrm{Emb}_{t_j}\}$, the most relevant frame and the most relevant transcription are retrieved independently:
\begin{equation}
i^* = \arg\max_i \ \mathrm{sim}(\mathrm{Emb}_{q_r}, \mathrm{Emb}_{f_i})
\end{equation}
\begin{equation}
j^* = \arg\max_j \ \mathrm{sim}(\mathrm{Emb}_{q_r}, \mathrm{Emb}_{t_j})
\end{equation}
where $\mathrm{sim}(\cdot, \cdot)$ denotes the cosine similarity:
\begin{equation}
\mathrm{sim}(\mathrm{Emb}_{q_r}, \mathbf{x}) = \frac{\mathrm{Emb}_{q_r} \cdot \mathbf{x}}{\|\mathrm{Emb}_{q_r}\| \|\mathbf{x}\|}
\end{equation}
with $\mathbf{x}$ representing either a frame embedding $\mathrm{Emb}_{f_i}$ or a transcription embedding $\mathrm{Emb}_{t_j}$.

To provide local context for the LVLM, we select a window of frames centered at $i^*$:
\begin{equation}
\mathcal{F}_{\mathrm{local}} = \{ \mathrm{Emb}_{f_{i^* - 1}}, \mathrm{Emb}_{f_{i^*}}, \mathrm{Emb}_{f_{i^*+1}} \}
\end{equation}

For transcriptions, we retrieve the top  $n$ most relevant segments based on similarity to $\mathrm{Emb}_{q_r}$:
\begin{equation}
J^\mathrm{trans}_{\mathrm{top}\text{  }n} = \underset{J \subset \{1,\ldots,M\}, |J|=n}{\arg\max} \sum_{j \in J} \mathrm{sim}(\mathrm{Emb}_{q_r}, \mathrm{Emb}_{t_j})
\end{equation}
The corresponding set of most relevant transcriptions is:
\begin{equation}
\mathcal{T}_{\mathrm{local}} = \{ T_j \mid j \in J^\mathrm{trans}_{\mathrm{top}\text{  }n} \}
\end{equation}

This approach ensures that the LVLM receives both the most contextually relevant frames (as a local temporal window) and the top  $n$ most pertinent transcript segments, along with the refined query $q_r$ and any other contextual cues to support robust answer generation.

The agent’s design is flexible: it can be adapted to select from a variety of function calls and can even be extended to interact with other agents or external sources, such as web search, depending on the needs of the application domain. For the experiments in this work, we focused on five core functions (see Table~\ref{tab:function_schema_for_the_pre_retrieval_agent}) to evaluate our methodology on the CinePile dataset. The agent selects which function to use based on its interpretation of the input query and the available function descriptions. For instance, when asked, "How many times did the character say 'fishing rod'?", the agent recognizes that this is a counting task and invokes the term frequency function to tally the occurrences of the phrase in the transcript.

Likewise, if a user query lacks specificity, the agent can employ the query expansion function to enrich the question with additional details from the global context. For example, "How does Yogi's mood change during the scene?" could be expanded to "How does Yogi (the bear wearing a green hat and white collar)'s mood change during the scene?", making the retrieval process more precise and targeted.

\begin{table*}[t]
\centering
\small
\caption{Comparison of accuracy (\%) across system variants and question types.}
\label{tab:comparison}
\begin{tabular}{
    l
    >{\centering\arraybackslash}p{2.2cm}
    >{\centering\arraybackslash}p{2.2cm}
    >{\centering\arraybackslash}p{2.2cm}
    >{\centering\arraybackslash}p{2.2cm}
    >{\centering\arraybackslash}p{2.0cm}
}

\toprule
\textbf{System Variant} & \textbf{Theme Exploration} & \textbf{Narrative and Plot Analysis} & \textbf{Character and Relationship Dynamics} & \textbf{Setting and Technical Analysis} & \textbf{Temporal} \\
\midrule
Baseline                          & 47.6          & 36.5          & 33.1          & 21.6          & 19.9 \\
+ Global Context Summary          & 51.3          & 37.4          & 35.1          & 23.2          & 23.1 \\
+ Pre  Retrieval Thinking Agent    & 55.0          & 43.2          & 38.4          & 26.2          & 24.9 \\
+ Rethink Module (AVATAAR)          & \textbf{55.6} & \textbf{44.7} & \textbf{39.2} & \textbf{26.6} & \textbf{25.5} \\
\bottomrule
\end{tabular}
\end{table*}
\vspace{1mm}
\vspace{2mm}
\subsubsection{Rethink Module} 
The Rethink Module is responsible for diagnosing and addressing cases where the LVLM’s response indicates insufficient information to answer the user’s query. It receives five types of input: (1) the user query, (2) expanded information from the Pre Retrieval Thinking Agent, (3) the global context summary, (4) the local context used (retrieved frames and transcriptions), and (5) the LVLM generated answer. Leveraging these inputs, the Rethink Module analyzes the reasoning process and identifies gaps or ambiguities that prevent the LVLM from producing a satisfactory answer. It then generates targeted instructions for the Pre Retrieval Thinking Agent to refine the query or retrieval process in subsequent iterations.

For example, if the user asks, “What is the correct sequence of events in the scene?” and the LVLM generated answer lacks temporal specificity, the Rethink Module may determine that the missing information is the exact timing of when “Ranger Smith confronts Yogi” occurs. In response, it issues an instruction such as “retrieve timestamps for the confrontation event,” prompting the Pre Retrieval Thinking Agent to invoke the temporal anchor extraction function. This iterative process continues for a predefined number of cycles or until a satisfactory answer is produced. By systematically diagnosing information gaps and orchestrating targeted retrieval strategies, the Rethink Module embodies an agentic RAG approach that enhances the robustness and accuracy of the overall framework.

In practice, we cap the number of rethink iterations at $x=3$ per question. The Rethink Module is prompted to review when we do not have enough information. This policy bounds the overhead of the iterative loop. The modest yet consistent gains from the Rethink Module therefore come at a controllable cost that can be tuned or disabled entirely in latency sensitive deployments.

\section{Evaluation}

\subsection{Video QA Dataset}

Selecting the evaluation corpus is pivotal for Video QA, as datasets differ markedly in their coverage of question categories, answer modalities, and annotation rigor. We surveyed widely used benchmarks: ActivityNetQA \cite{activitynetqa}, NExT  QA \cite{NExT-QA}, and CinePile \cite{rawal2024cinepile} against three criteria:
\begin{enumerate}
\item Breadth across question types (temporal, attribute, narrative, thematic);
\item Answer modality (open-ended vs.\ multiple choice);
\item Practicality and reliability of automatic validation.
\end{enumerate}

Existing benchmarks such as \textbf{ActivityNetQA} and \textbf{NExT-QA} emphasize open-ended activity recognition and causal/temporal reasoning, but their free-form answers make automatic evaluation noisier and narrow their question coverage for narrative and thematic analysis.

By contrast, \textbf{CinePile} offers a balanced distribution of temporal, attribute, narrative, and thematic questions, all delivered as multiple choice. This design substantially reduces ambiguity during validation and supports a more reliable, objective evaluation. CinePile also features diverse video content and careful annotation; accordingly, we adopt it for our Video QA benchmarking. For definitions and examples of each question type, see Table~\ref{tab:question  types}.

\subsection{Evaluation Metrics}

We evaluate the different framework designs using two standard metrics: accuracy and F1 score. Accuracy captures the overall proportion of correct predictions, providing a straightforward measure of performance. The F1 score balances precision and recall, making it especially useful when the dataset is uneven across classes. Together, these metrics provide a reliable view of both general correctness and performance under imbalance, and we report both in all experiments.

\begin{figure*}[t]
\centering
\includegraphics[width=0.75\textwidth]{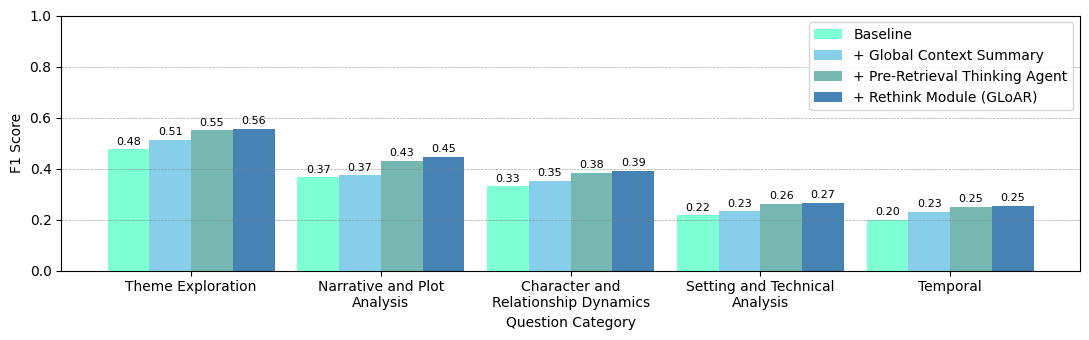}
\caption{F1 Score by System Variant and Category}
\label{fig:AVATAAR_f1}
\end{figure*}


\paragraph*{Evaluation protocol}
CinePile is a multiple choice benchmark: each question $q$ is paired with five candidate answers and we add one more answer option making total 6 choices $\{a_1,\dots,a_6\}$. At inference time, we prompt the LVLM to output only a single option label (\texttt{1}, \texttt{2}, \texttt{3}, \texttt{4}, \texttt{5} , or \texttt{6}), which we deterministically map to the corresponding ans index. Accuracy is computed as the proportion of questions where the predicted index matches $y^\star$. F1 is computed over the induced six way classification problem and macro-averaged across answer options. The model is constrained to select from the finite choice set, there is no ambiguity in mapping generated outputs to ground truth labels.

\subsection{Component Impacts}

To assess the effectiveness of each component in the AVATAAR framework, we compare four system variants: (1) the baseline video chat framework without any AVATAAR components, (2) the baseline enhanced with the global context summary, (3) the baseline further augmented with the Pre Retrieval Thinking Agent, and (4) the complete AVATAAR framework, which also includes the Rethink Module. All models are evaluated on the same test split of the CinePile dataset, with accuracy serving as the primary metric. All experiments use the same models and model configurations to ensure a fair comparison.

Adding the Pre Retrieval Thinking Agent yields further accuracy gains across all categories, such as increases of 3.7\% (from 51.3\% to 55.0\%) for Theme Exploration and 5.6\% (from 37.4\% to 43.0\%) for Narrative. Although these improvements are moderate, they demonstrate that dynamic function selection and query refinement enhance AVATAAR’s ability to handle tasks requiring more complex reasoning. 

Table~\ref{tab:comparison} shows that adding global context yields consistent gains over the baseline, particularly for Temporal and Theme Exploration questions. The Pre Retrieval Thinking Agent further improves all categories, and the full AVATAAR system with the Rethink Module brings modest but consistent additional improvements (0.4 – 1.7 percentage points). Figure~\ref{fig:AVATAAR_f1} confirms the same trend in terms of F1 scores across question types.

To give a broader evaluation of our different system setups, we also looked at F1 scores, as you can see in Figure~\ref{fig:AVATAAR_f1}. Using F1 scores helps paint a fuller picture of how the systems perform, especially when there are class imbalances or when answers are not completely correct. Our findings show that AVATAAR generally does better than the baseline models, particularly when it comes to Theme Exploration, a category where understanding themes and emotional nuances is key. Interestingly, even though Theme Exploration has the least number of questions, just 189 per Table~\ref{tab:question  types}, we saw quite a bit of improvement there, which suggests AVATAAR can handle abstract concepts pretty well.

On the flip side, Figure~\ref{fig:AVATAAR_f1} also highlights how Temporal questions, crucial for understanding sequences in videos (with 683 questions according to the table), remain tricky. Still, the F1 scores back up the accuracy findings, giving us confidence in the results. These tests show that the different parts of AVATAAR, like how it models global context and refines queries, play their roles in boosting video Q\&A. Our setup does show improvements over older methods, especially for Theme Exploration and Narrative questions, with certain gains around 8.0\%.

Although recent models like Gemini 1.5 Pro and GPT  4o have achieved higher performance than our framework on the CinePile benchmark (as reported in the CinePile paper~\cite{rawal2024cinepile}), our goal is not to introduce the absolute best performing video understanding model. Instead, we aim to present a flexible framework that can be deployed or built with any multimodal models, including those that may become available in the future. AVATAAR is specifically designed for enterprise use on proprietary infrastructure, which allows organizations to avoid reliance on external APIs whose internal mechanisms may be opaque or outside their control.

\section{Conclusion}

In this work, we introduced AVATAAR, a modular and comprehensible framework designed for long-form video question answering that prioritizes retrieval  augmented, agentic reasoning over simply scaling up models. By clearly distinguishing global summarization from local evidence retrieval and integrating an iterative feedback loop between the Pre Retrieval Thinking Agent and the Rethink Module, AVATAAR enables more context aware and precise responses across a wide array of complex video queries. Our evaluation of the CinePile dataset shows notable improvements, especially in addressing intricate spatio-temporal and narrative questions. AVATAAR stands out as a transparent and flexible alternative to proprietary video QA systems, offering enterprises greater adaptability and control over their applications. We are confident that the architectural innovations provided by AVATAAR establish a robust foundation for the future of multimodal video understanding. This framework is poised to facilitate not just efficient adaptation to diverse domains but also to ensure strong performance in real-world, content heavy scenarios. Its design supports further innovation, making it a valuable tool for ongoing developments in the field.



\bibliographystyle{IEEEtran}
\bibliography{IEEEexample}

\end{document}